\documentclass[11pt]{article}
\usepackage[utf8]{inputenc}
\usepackage[T1]{fontenc}
\usepackage{graphicx}
\usepackage{grffile}
\usepackage{longtable}
\usepackage{wrapfig}
\usepackage{rotating}
\usepackage[normalem]{ulem}
\usepackage{amsmath}
\usepackage{textcomp}
\usepackage{amssymb}
\usepackage{capt-of}
\usepackage{hyperref}
\usepackage[usenames,dvipsnames]{xcolor}
\usepackage{geometry}
\usepackage[activate={true,nocompatibility},final,tracking=true,kerning=true,spacing=nonfrench,factor=1100,stretch=10,shrink=10]{microtype}

\usepackage{booktabs}
\usepackage{authblk}
\usepackage{mathtools}
\usepackage{amssymb}
\usepackage{natbib}
\usepackage{physics}
\usepackage{subfigure}
\usepackage{times}
\usepackage{tikz}

\DeclareMathOperator*{\argmin}{argmin}

\DeclareMathOperator*{\sign}{sign}
\newcommand\pred[1]{\overline{#1}}
\newcommand\given{\:\vert\:}

\usepackage[accepted]{icml2017}

\author{Zhitao Gong}
\date{\today}
\title{}
\hypersetup{
      pdfauthor={Zhitao Gong},
      pdftitle={},
      pdfkeywords={},
      pdfsubject={},
      pdfcreator={Emacs 25.1.1 (Org mode 9.0.5)},
      pdflang={English},
      bookmarks=true,
      unicode=true,
      pdftoolbar=true,
      pdfmenubar=true,
      pdffitwindow=false,
      pdfstartview={FitH},
      pdfnewwindow=true,
      colorlinks=true,
      linkcolor=Maroon,
      citecolor=ForestGreen,
      filecolor=Mulberry,
      urlcolor=MidnightBlue}
\begin{document}


\twocolumn[
\icmltitle{Adversarial and Clean Data Are Not Twins}

\begin{icmlauthorlist}
\icmlauthor{Zhitao Gong}{au}
\icmlauthor{Wenlu Wang}{au}
\icmlauthor{Wei-Shinn Ku}{au}
\end{icmlauthorlist}

\icmlaffiliation{au}{Auburn University, Auburn, AL}

\icmlcorrespondingauthor{Zhitao Gong}{gong@auburn.edu}


\icmlkeywords{adversarial, deep neural network}

\vskip 0.3in
]

\printAffiliationsAndNotice{}

\begin{abstract}

Adversarial attack has cast a shadow on the massive success of deep
neural networks.  Despite being almost visually identical to the clean
data, the adversarial images can fool deep neural networks into wrong
predictions with very high confidence.  In this paper, however, we
show that we can build a simple binary classifier separating the
adversarial apart from the clean data with accuracy over 99\%.  We also
empirically show that the binary classifier is robust to a
second-round adversarial attack.  In other words, it is difficult to
disguise adversarial samples to bypass the binary classifier.  Further
more, we empirically investigate the generalization limitation which
lingers on all current defensive methods, including the binary
classifier approach.  And we hypothesize that this is the result of
intrinsic property of adversarial crafting algorithms.
\end{abstract}

\section{Introduction}
\label{sec:introduction}
\begin{figure*}
\centering
\includegraphics[width=.9\linewidth]{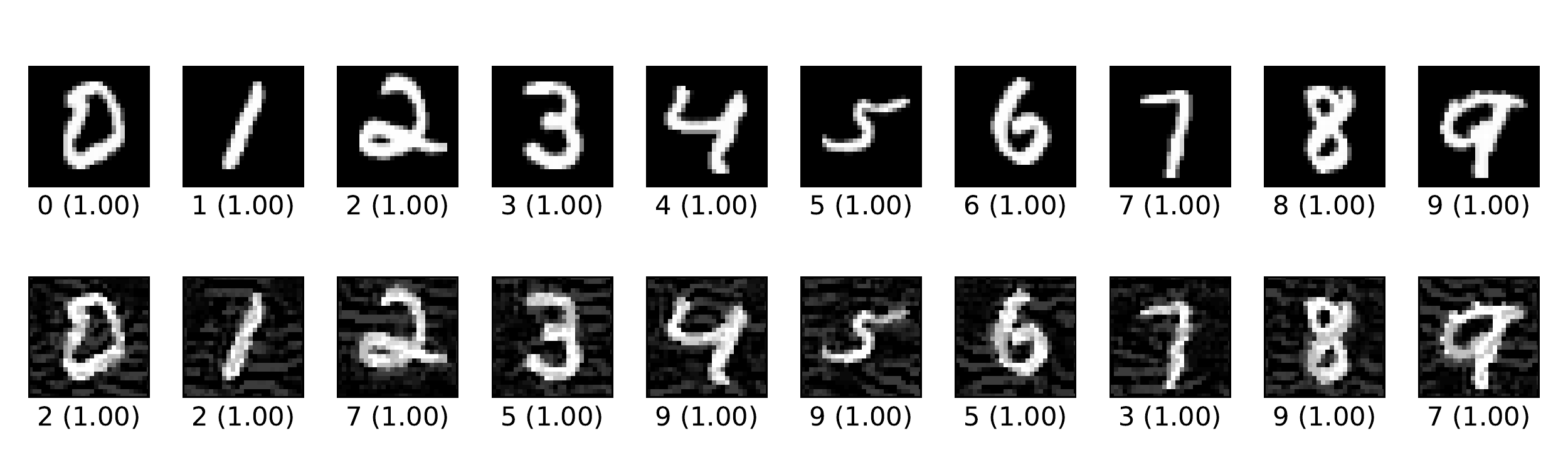}
\caption{\label{fig:adv-example}
The adversarial images (second row) are generated from the first row via iterative FGSM.  The label of each image is shown below with prediction probability in parenthesis.  Our model achieves less then 1\% error rate on the clean data.}
\end{figure*}

Deep neural networks have been successfully adopted to many life
critical areas, e.g., skin cancer detection
\cite{esteva2017-dermatologist}, auto-driving \cite{santana2016-learning},
traffic sign classification \cite{ciresan2012-multi}, etc.  A recent
study \cite{szegedy2013-intriguing}, however, discovered that deep
neural networks are susceptible to adversarial images.  Figure
\ref{fig:adv-example} shows an example of adversarial images generated
via fast gradient sign method
\cite{kurakin2016-adversarial,kurakin2016-adversarial-1} on MNIST.  As
we can see that although the adversarial and original clean images are
almost identical from the perspective of human beings, the deep neural
network will produce wrong predictions with very high confidence.
Similar techniques can easily fool the image system into mistaking a
stop sign for a yield sign, a dog for a automobile, for example.  When
leveraged by malicious users, these adversarial images pose a great
threat to the deep neural network systems.

Although adversarial and clean images appear visually indiscernible,
their subtle differences can successfully fool the deep neural
networks.  This means that deep neural networks are sensitive to these
subtle differences.  So an intuitively question to ask is: can we
leverage these subtle differences to distinguish between adversarial
and clean images?  Our experiment suggests the answer is positive.  In
this paper we demonstrate that a simple binary classifier can separate
the adversarial from the original clean images with very high accuracy
(over 99\%).  However, we also show that the binary classifier approach
suffers from the generalization limitation, i.e., it is sensitive 1)
to a hyper-parameter used in crafting adversarial dataset, and 2) to
different adversarial crafting algorithms.  In addition to that, we
also discovered that this limitation is also shared among other
proposed methods against adversarial attacking, e.g., defensive
retraining \cite{huang2015-learning,kurakin2016-adversarial-1},
knowledge distillation \cite{papernot2015-distillation}, etc.  We
empirically investigate the limitation and propose the hypothesis that
the adversarial and original dataset are, in effect, two completely
\emph{different} datasets, despite being visually similar.

This article is organized as follows.  In Section
\ref{sec:related-work}, we give an overview of the current research in
adversarial attack and defense, with a focus on deep neural networks.
Then, it is followed by a brief summary of the state-of-the-art
adversarial crafting algorithms in Section
\ref{sec:crafting-adversarials}.  Section \ref{sec:experiment} presents
our experiment results and detailed discussions.  And we conclude in
Section \ref{sec:conclusion}.

\section{Related Work}
\label{sec:related-work}
The adversarial image attack on deep neural networks was first
investigated in \cite{szegedy2013-intriguing}.  The authors discovered
that when added some imperceptible carefully chosen noise, an image
may be wrongly classified with high confidence by a well-trained deep
neural network.  They also proposed an adversarial crafting algorithm
based on optimization.  We will briefly summarize it in section
\ref{sec:crafting-adversarials}.  They also proposed the hypothesis that
the adversarial samples exist as a result of the high nonlinearity of
deep neural network models.

However, \cite{goodfellow2014-explaining} proposed a counter-intuitive
hypothesis explaining the cause of adversarial samples.  They argued
that adversarial samples are caused by the models being too \emph{linear},
rather than \emph{nonlinear}.  They proposed two adversarial crafting
algorithms based on this hypothesis, i.e., fast gradient sign method
(FGSM) and least-likely class method (LLCM)
\cite{goodfellow2014-explaining}.  The least-likely class method is
later generalized to target class gradient sign method (TGSM) in
\cite{kurakin2016-adversarial}.

\cite{papernot2015-limitations} proposed another gradient based
adversarial algorithm, the Jacobian-based saliency map approach (JSMA)
which can successfully alter the label of an image to any desired
category.

The adversarial images have been shown to be transferable among deep
neural networks \cite{szegedy2013-intriguing,kurakin2016-adversarial}.
This poses a great threat to current learning systems in that the
attacker needs not the knowledge of the target system.  Instead, the
attacker can train a different model to create adversarial samples
which are still effective for the target deep neural networks.  What's
worse, \cite{papernot2016-transferability} has shown that adversarial
samples are even transferable among different machine learning
techniques, e.g., deep neural networks, support vector machine,
decision tree, logistic regression, etc.

Small steps have been made towards the defense of adversarial images.
\cite{kurakin2016-adversarial} shows that some image transformations,
e.g., Gaussian noise, Gaussian filter, JPEG compression, etc., can
effectively recover over 80\% of the adversarial images.  However, in
our experiment, the image transformation defense does not perform well
on images with low resolution, e.g., MNIST.  Knowledge distillation is
also shown to be an effective method against most adversarial images
\cite{papernot2015-distillation}.  The restrictions of defensive
knowledge distillation are 1) that it only applies to models that
produce categorical probabilities, and 2) that it needs model
retraining.  Adversarial training
\cite{kurakin2016-adversarial-1,huang2015-learning} was also shown to
greatly enhance the model robustness to adversarials.  However, as
discussed in Section \ref{subsec:generalization-limitation}, defensive
distillation and adversarial training suffers from, what we call, the
generalization limitations.  Our experiment suggests this seems to be
an intrinsic property of adversarial datasets.

\section{Crafting Adversarials}
\label{sec:crafting-adversarials}
The are mainly two categories of algorithms to generate adversarial
samples, model independent and model dependent.  We briefly summarize
these two classes of methods in this section.

By conventions, we use \(X\) to represent input image set (usually a
3-dimension tensor), and \(Y\) to represent the label set, usually
one-hot encoded.  Lowercase represents an individual data sample,
e.g., \(x\) for one input image.  Subscript to data samples denotes
one of its elements, e.g., \(x_i\) denotes one pixel in the image,
\(y_i\) denotes probability for the \(i\)-th target class.  \(f\)
denotes the model, \(\theta\) the model parameter, \(J\) the loss
function.  We use the superscript \emph{adv} to denote adversarial related
variables, e.g., \(x^{adv}\) for one adversarial image.  \(\delta x\)
denotes the adversarial noise for one image, i.e., \(x^{adv} = x +
\delta x\).  For clarity, we also include the model used to craft the
adversarial samples where necessary, e.g., \(x^{adv(f_1)}\) denotes
the adversarial samples created with model \(f_1\).  \(\mathbb{D}\)
denotes the image value domain, usually \([0, 1]\) or \([0, 255]\).
And \(\epsilon\) is a scalar controlling the scale of the adversarial
noise, another hyper-parameter to choose.

\subsection{Model Independent Method}
\label{sec:org5632935}

A box-constrained minimization algorithm based on L-BFGS was the first
algorithm proposed to generate adversarial data
\cite{szegedy2013-intriguing}.  Concretely we want to find the smallest
(in the sense of \(L^2\)-norm) noise \(\delta x\) such that the
adversarial image belongs to a different category, i.e.,
\(f(x^{adv})\neq f(x)\).
\begin{equation} \label{eq:guided-walk}
  \begin{split}
    \delta x &= \argmin_r c\norm{r}_\infty + J(x+r, y^{adv})\\
    &\text{ s.t. } x+r\in \mathbb{D}
  \end{split}
\end{equation}

\subsection{Model Dependent Methods}
\label{sec:org38c3a5e}

There are mainly three methods that rely on model gradient, i.e., fast
gradient sign method (FGSM) \cite{kurakin2016-adversarial}, target class
method \cite{kurakin2016-adversarial,kurakin2016-adversarial-1} (TGSM)
and Jacobian-based saliency map approach (JSMA)
\cite{papernot2015-limitations}.  We will see in Section
\ref{sec:experiment} that despite that they all produce highly
disguising adversarials, FGSM and TGSM produce \emph{compatible}
adversarial datasets which are complete \emph{different} from adversarials
generated via JSMA.

\subsubsection*{Fast Gradient Sign Method (FGSM)}
\label{sec:orgd627680}

FGSM tries to modify the input towards the direction where \(J\)
increases, i.e., \(\dv*{J(x, y^{adv})}{x}\), as shown in Equation
\ref{eq:fgsm}.
\begin{equation} \label{eq:fgsm}
  \delta x = \epsilon\sign\left(\dv{J(x, \pred{y})}{x}\right)
\end{equation}

Originally \cite{kurakin2016-adversarial} proposes to generate
adversarial samples by using the true label i.e., \(y^{adv} =
y^{true}\), which has been shown to suffer from the label leaking
problem \cite{kurakin2016-adversarial-1}.  Instead of true labels,
\cite{kurakin2016-adversarial-1} proposes to use the \emph{predicted} label,
i.e., \(\pred{y} = f(x)\), to generate adversarial examples.

This method can also be used iteratively as shown in Equation
\ref{eq:fgsm-iter}.  Iterative FGSM has much higher success rate than
the one-step FGSM.  However, the iterative version is less robust to
image transformation \cite{kurakin2016-adversarial}.
\begin{equation} \label{eq:fgsm-iter}
  \begin{split}
    x^{adv}_{k+1} &= x^{adv}_k + \epsilon\sign\left(\dv{J(x^{adv}_k, \pred{y_k})}{x}\right)\\
    x^{adv}_0 &= x\\
    \pred{y_k} &= f(x^{adv}_k)
  \end{split}
\end{equation}

\subsubsection*{Target Class Gradient Sign Method (TGSM)}
\label{sec:orge23b357}

This method tries to modify the input towards the direction where
\(p(y^{adv}\given x)\) increases.
\begin{equation} \label{eq:tcm}
    \delta x = -\epsilon\sign\left(\dv{J(x, y^{adv})}{x}\right)
\end{equation}

Originally this method was proposed as the least-likely class method
\cite{kurakin2016-adversarial} where \(y^{adv}\) was chosen as the
least-likely class predicted by the model as shown in Equation
\ref{eq:llcm-y}.
\begin{equation} \label{eq:llcm-y}
  y^{adv} = \text{OneHotEncode}\left(\argmin f(x)\right)
\end{equation}

And it was extended to a more general case where \(y^{adv}\) could be
any desired target class \cite{kurakin2016-adversarial-1}.

\begin{table*}[htbp]
  \caption{\label{tbl:accuracy-summary}
    Accuracy on adversarial samples generated with FGSM/TGSM.}
  \centering
  \begin{tabular}{lcrrcrrrr}
    \toprule
    & \phantom{a} & \multicolumn{2}{c}{\(f_1\)} & \phantom{a} & \multicolumn{4}{c}{\(f_2\)} \\
    \cmidrule{3-4} \cmidrule{6-9}
    Dataset && \(X_{test}\) & \(X^{adv(f_1)}_{test}\) && \(X_{test}\) & \(X^{adv(f_1)}_{test}\) & \(\{X_{test}\}^{adv(f_2)}\) & \(\{X^{adv(f_1)}_{test}\}^{adv(f_2)}\) \\
    \midrule
    MNIST && 0.9914 & 0.0213 && 1.00 & 1.00 & 0.00 & 1.00\\
    CIFAR10 && 0.8279 & 0.1500 && 0.99 & 1.00 & 0.01 & 1.00\\
    SVHN && 0.9378 & 0.2453 && 1.00 & 1.00 & 0.00 & 1.00\\
    \bottomrule
  \end{tabular}
\end{table*}

\subsubsection*{Jacobian-based Saliency Map Approach (JSMA)}
\label{sec:org6e474fd}

Similar to the target class method, JSMA \cite{papernot2015-limitations}
allows to specify the desired target class.  However, instead of
adding noise to the whole input, JSMA changes only one pixel at a
time.  A \emph{saliency score} is calculated for each pixel and pixel with
the highest score is chosen to be perturbed.
\begin{equation} \label{eq:jsma-saliency}
  \begin{split}
    s(x_i) &= \begin{cases}
      0 & \text{ if } s_t < 0 \text{ or } s_o > 0\\
      s_t\abs{s_o} & \text{ otherwise}
    \end{cases}\\
    s_t &= \pdv{y_t}{x_i}\qquad s_o = \sum_{j\neq t}\pdv{y_j}{x_i}
  \end{split}
\end{equation}

Concretely, \(s_t\) is the Jacobian value of the desired target class
\(y_t\) w.r.t an individual pixel, \(s_o\) is the sum of Jacobian
values of all non-target class.  Intuitively, saliency score indicates
the sensitivity of each output class w.r.t each individual pixel.  And
we want to perturb the pixel towards the direction where \(p(y_t\given
x)\) increases the most.

\section{Experiment}
\label{sec:experiment}
Generally, we follow the steps below to test the effectiveness and
limitation of the binary classifier approach.

\begin{enumerate}
\item Train a deep neural network \(f_1\) on the original clean training
data \(X_{train}\), and craft adversarial dataset from the original
clean data, \(X_{train}\to X^{adv(f_1)}_{train}\), \(X_{test}\to
   X^{adv(f_1)}_{test}\).  \(f_1\) is used to generate the attacking
adversarial dataset which we want to filter out.
\item Train a binary classifier \(f_2\) on the combined (shuffled)
training data \(\{X_{train}, X^{adv(f_1)}_{train}\}\), where
\(X_{train}\) is labeled 0 and \(X^{adv(f_1)}_{train}\) labeled 1.
\item Test the accuracy of \(f_2\) on \(X_{test}\) and
\(X^{adv(f_1)}_{test}\), respectively.
\item Construct second-round adversarial test data, \(\{X_{test},
   X^{adv(f_1)}_{test}\}\to \{X_{test},
   X^{adv(f_1)}_{test}\}^{adv(f_2)}\) and test \(f_2\) accuracy on
this new adversarial dataset.  Concretely, we want to test whether
we could find adversarial samples 1) that can successfully bypass
the binary classifier \(f_2\), and 2) that can still fool the
target model \(f_1\) if they bypass the binary classifier.  Since
adversarial datasets are shown to be transferable among different
machine learning techniques \cite{papernot2016-transferability}, the
binary classifier approach will be seriously flawed if \(f_2\)
failed this second-round attacking test.
\end{enumerate}

The code to reproduce our experiment are available
\url{https://github.com/gongzhitaao/adversarial-classifier}.

\subsection{Efficiency and Robustness of the Classifier}
\label{sec:orgd914749}

We evaluate the binary classifier approach on MNIST, CIFAR10, and SVHN
datasets.  Of all the datasets, the binary classifier achieved
accuracy over 99\% and was shown to be robust to a second-round
adversarial attack.  The results are summarized in Table
\ref{tbl:accuracy-summary}.  Each column denotes the model accuracy on
the corresponding dataset.  The direct conclusions from Table
\ref{tbl:accuracy-summary} are summarized as follows.
\begin{enumerate}
\item Accuracy on \(X_{test}\) and \(X^{adv(f_1)}_{test}\) suggests that
the binary classifier is very effective at separating adversarial
from clean dataset.  Actually during our experiment, the accuracy
on \(X_{test}\) is always near 1, while the accuracy on
\(X^{adv(f_1)}_{test}\) is either near 1 (successful) or near 0
(unsuccessful).  Which means that the classifier either
successfully detects the subtle difference completely or fails
completely.  We did not observe any values in between.
\item Accuracy on \(\{X^{adv(f_1)}_{test}\}^{adv(f_2)}\) suggests that we
were not successful in disguising adversarial samples to bypass the
the classifier.  In other words, the binary classifier approach is
robust to a second-round adversarial attack.
\item Accuracy on \(\{X_{test}\}^{adv(f_2)}\) suggests that in case of
the second-round attack, the binary classifier has very high false
negative.  In other words, it tends to recognize them all as
adversarials.  This, does not pose a problem in our opinion.  Since
our main focus is to block adversarial samples.
\end{enumerate}

\subsection{Generalization Limitation}
\label{subsec:generalization-limitation}
Before we conclude too optimistic about the binary classifier approach
performance, however, we discover that it suffers from the
\emph{generalization limitation}.
\begin{enumerate}
\item When trained to recognize adversarial dataset generated via
FGSM/TGSM, the binary classifier is sensitive to the
hyper-parameter \(\epsilon\).
\item The binary classifier is also sensitive to the adversarial crafting
algorithm.
\end{enumerate}

In out experiment, the aforementioned limitations also apply to
adversarial training \cite{kurakin2016-adversarial-1,huang2015-learning}
and defensive distillation \cite{papernot2015-distillation}.

\subsubsection*{Sensitivity to \(\epsilon\)}
\label{sec:org09dbb1e}

Table \ref{tbl:eps-sensitivity-cifar10} summarizes our tests on CIFAR10.
For brevity, we use \(\eval{f_2}_{\epsilon=\epsilon_0}\) to denote
that the classifier \(f_2\) is trained on adversarial data generated
on \(f_1\) with \(\epsilon=\epsilon_0\).  The binary classifier is
trained on mixed clean data and adversarial dataset which is generated
via FGSM with \(\epsilon=0.03\).  Then we re-generate adversarial
dataset via FGSM/TGSM with different \(\epsilon\) values.

\begin{table}[htbp]
  \caption{\label{tbl:eps-sensitivity-cifar10}
    \(\epsilon\) sensitivity on CIFAR10}
  \centering
  \begin{tabular}{lcll}
    \toprule
    & \phantom{a} & \multicolumn{2}{c}{\(\eval{f_2}_{\epsilon=0.03}\)} \\
    \cmidrule{3-4}
    \(\epsilon\) && \(X_{test}\) & \(X^{adv(f_1)}_{test}\)\\
    \midrule
    0.3 && 0.9996 & 1.0000\\
    0.1 && 0.9996 & 1.0000\\
    0.03 && 0.9996 & 0.9997\\
    0.01 && 0.9996 & \textbf{0.0030}\\
    \bottomrule
  \end{tabular}
\end{table}

As shown in Table \ref{tbl:eps-sensitivity-cifar10},
\(\eval{f_2}_{\epsilon=\epsilon_0}\) can correctly filter out
adversarial dataset generated with \(\epsilon\geq\epsilon_0\), but
fails when adversarial data are generated with
\(\epsilon<\epsilon_1\).  Results on MNIST and SVHN are similar.  This
phenomenon was also observed in defensive retraining
\cite{kurakin2016-adversarial-1}.  To overcome this issue, they proposed
to use mixed \(\epsilon\) values to generate the adversarial datasets.
However, Table \ref{tbl:eps-sensitivity-cifar10} suggests that
adversarial datasets generated with smaller \(\epsilon\) are
\emph{superset} of those generated with larger \(\epsilon\).  This
hypothesis could be well explained by the linearity hypothesis
\cite{kurakin2016-adversarial,warde-farley2016-adversarial}.  The same
conclusion also applies to adversarial training.  In our experiment,
the results of defensive retraining are similar to the binary
classifier approach.

\subsubsection*{Disparity among Adversarial Samples}
\label{sec:orge9aa2b5}

\begin{figure*}
\centering
\includegraphics[width=.9\linewidth]{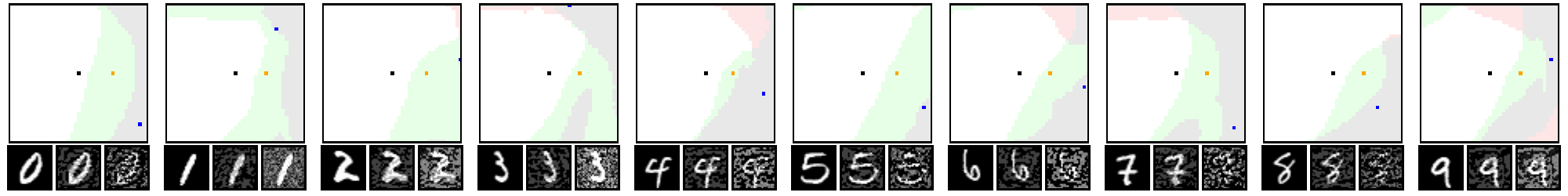}
\caption{\label{fig:adv-training-not-working}
Adversarial training \cite{huang2015-learning,kurakin2016-adversarial-1} does not work.  This is a church window plot \cite{warde-farley2016-adversarial}.  Each pixel \((i, j)\) (row index and column index pair) represents a data point \(\tilde{x}\) in the input space and \(\tilde{x} = x + \vb{h}\epsilon_j + \vb{v}\epsilon_i\), where \(\vb{h}\) is the direction computed by FGSM and \(\vb{v}\) is a random direction orthogonal to \(\vb{h}\).  The \(\epsilon\) ranges from \([-0.5, 0.5]\) and \(\epsilon_{(\cdot)}\) is the interpolated value in between.  The central black dot \tikz[baseline=-0.5ex]{\draw[fill=black] (0,0) circle (0.3ex)} represents the original data point \(x\), the orange dot (on the right of the center dot) \tikz[baseline=-0.5ex]{\draw[fill=orange,draw=none] (0,0) circle (0.3ex)} represents the last adversarial sample created from \(x\) via FGSM that is used in the adversarial training and the blue dot \tikz[baseline=-0.5ex]{\draw[fill=blue,draw=none] (0,0) circle (0.3ex)} represents a random adversarial sample created from \(x\) that cannot be recognized with adversarial training. The three digits below each image, from left to right, are the data samples that correspond to the black dot, orange dot and blue dot, respectively.  \tikz[baseline=0.5ex]{\draw (0,0) rectangle (2.5ex,2.5ex)} ( \tikz[baseline=0.5ex]{\draw[fill=black,opacity=0.1] (0,0) rectangle (2.5ex,2.5ex)} ) represents the data samples that are always correctly (incorrectly) recognized by the model.  \tikz[baseline=0.5ex]{\draw[fill=red,opacity=0.1] (0,0) rectangle (2.5ex,2.5ex)} represents the adversarial samples that can be correctly recognized without adversarial training only.  And \tikz[baseline=0.5ex]{\draw[fill=green,opacity=0.1] (0,0) rectangle (2.5ex,2.5ex)} represents the data points that were correctly recognized with adversarial training only, i.e., the side effect of adversarial training.}
\end{figure*}

In our experiment, we also discovered that the binary classifier is
also sensitive to the algorithms used to generate the adversarial
datasets.

Specifically, the binary classifier trained on FGSM adversarial
dataset achieves good accuracy (over 99\%) on FGSM adversarial dataset,
but not on adversarial generated via JSMA, and vise versa.  However,
when binary classifier is trained on a mixed adversarial dataset from
FGSM and JSMA, it performs well (with accuracy over 99\%) on both
datasets.  This suggests that FGSM and JSMA generate adversarial
datasets that are \emph{far away} from each other.  It is too vague without
defining precisely what is \emph{being far away}.  In our opinion, they are
\emph{far away} in the same way that CIFAR10 is \emph{far away} from SVHN.  A
well-trained model on CIFAR10 will perform poorly on SVHN, and vise
versa.  However, a well-trained model on the the mixed dataset of
CIFAR10 and SVHN will perform just as well, if not better, on both
datasets, as if it is trained solely on one dataset.

The adversarial datasets generated via FGSM and TGSM are, however,
\emph{compatible} with each other.  In other words, the classifier trained
on one adversarial datasets performs well on adversarials from the
other algorithm.  They are compatible in the same way that training
set and test set are compatible.  Usually we expect a model, when
properly trained, should generalize well to the unseen data from the
same distribution, e.g., the test dataset.

In effect, it is not just FGSM and JSMA are incompatible.  We can
generate adversarial data samples by a linear combination of the
direction computed by FGSM and another random orthogonal direction, as
illustrated in a church plot \cite{warde-farley2016-adversarial} Figure
\ref{fig:adv-training-not-working}.  Figure
\ref{fig:adv-training-not-working} visually shows the effect of
adversarial training \cite{kurakin2016-adversarial-1}.  Each image
represents adversarial samples generated from \emph{one} data sample, which
is represented as a black dot in the center of each image, the last
adversarial sample used in adversarial training is represented as an
orange dot (on the right of black dot, i.e., in the direction computed
by FGSM).  The green area represents the adversarial samples that
cannot be correctly recognized without adversarial training but can be
correctly recognized with adversarial training.  The red area
represents data samples that can be correctly recognized without
adversarial training but cannot be correctly recognized with
adversarial training.  In other words, it represents the side effect
of adversarial training, i.e., slightly reducing the model accuracy.
The white (gray) area represents the data samples that are always
correctly (incorrectly) recognized with or without adversarial
training.

As we can see from Figure \ref{fig:adv-training-not-working},
adversarial training does make the model more robust against the
adversarial sample (and adversarial samples around it to some extent)
used for training (green area).  However, it does not rule out all
adversarials.  There are still adversarial samples (gray area) that
are not affected by the adversarial training.  Further more, we could
observe that the green area largely distributes along the horizontal
direction, i.e., the FGSM direction.  In \cite{nguyen2014-deep}, they
observed similar results for fooling images.  In their experiment,
adversarial training with fooling images, deep neural network models
are more robust against a limited set of fooling images.  However they
can still be fooled by other fooling images easily.

\section{Conclusion}
\label{sec:conclusion}
We show in this paper that the binary classifier is a simple yet
effective and robust way to separating adversarial from the original
clean images.  Its advantage over defensive retraining and
distillation is that it serves as a preprocessing step without
assumptions about the model it protects.  Besides, it can be readily
deployed without any modification of the underlying systems.  However,
as we empirically showed in the experiment, the binary classifier
approach, defensive retraining and distillation all suffer from the
generalization limitation.  For future work, we plan to extend our
current work in two directions.  First, we want to investigate the
disparity between different adversarial crafting methods and its
effect on the generated adversarial space.  Second, we will also
carefully examine the cause of adversarial samples since intuitively
the linear hypothesis does not seem right to us.

\bibliographystyle{icml2017}
\bibliography{/home/gongzhitaao/Dropbox/bibliography/nn}
\end{document}